\pdfoutput=1
\pdfoutput=1
\pdfoutput=1
\pdfoutput=1
\pdfoutput=1
\documentclass[10pt,twocolumn,letterpaper]{article}
\usepackage{iccv}
\usepackage{times}
\usepackage{epsfig}
\usepackage{graphicx}
\usepackage{amsmath}
\usepackage{amssymb}
\usepackage{multirow}
\usepackage{bbm}
\usepackage{subcaption}
\usepackage{algorithm}
\usepackage{algorithmic}
\usepackage{bm}
\usepackage{float}
\usepackage{footnote}

\usepackage[breaklinks=true,bookmarks=false]{hyperref}

\iccvfinalcopy 

\def\iccvPaperID{1517} 

\ificcvfinal\fi

\begin{document}

\title{Learning Object-Specific Distance From a Monocular Image}

\author{Jing Zhu$^{1,2,3}$ \hspace{0.5cm} Yi Fang$^{1,2,3}$\thanks{indicates corresponding author.} \hspace{0.5cm} Husam Abu-Haimed$^{4}$ \hspace{0.5cm} Kuo-Chin Lien$^{4}$ \hspace{0.5cm} Dongdong Fu$^{4}$ \hspace{0.5cm} Junli Gu$^{4}$ \vspace{0.2cm}\\
$^{1}$NYU Multimedia and Visual Computing Lab, USA \\
$^{2}$New York University, USA \\
$^{3}$New York University Abu Dhabi, UAE\\
$^{4}$XMotors.ai \\
{\tt\small {\{jingzhu, yfang\}@nyu.edu \hspace{0.5cm} husam.abu.haimed@gmail.com \hspace{0.5cm} \{kuochin, dongdong, junli\}@xmotors.ai}}}

\maketitle
\ificcvfinal\thispagestyle{empty}\fi

\begin{abstract}
 Environment perception, including object detection and distance estimation, is one of the most crucial tasks for autonomous driving. Many attentions have been paid on the object detection task, but distance estimation only arouse few interests in the computer vision community. Observing that the traditional inverse perspective mapping algorithm performs poorly for objects far away from the camera or on the curved road, in this paper, we address the challenging distance estimation problem by developing the first end-to-end learning-based model to directly predict distances for given objects in the images. Besides the introduction of a learning-based base model, we further design an enhanced model with a keypoint regressor, where a projection loss is defined to enforce a better distance estimation, especially for objects close to the camera. To facilitate the research on this task, we construct the extented KITTI and nuScenes (mini)  object detection datasets with a distance for each object. Our experiments demonstrate that our proposed methods outperform alternative approaches (e.g., the traditional IPM, SVR) on object-specific distance estimation, particularly for the challenging cases that objects are on a curved road. Moreover, the performance margin implies the effectiveness of our enhanced method. 
\end{abstract}


\section{Introduction}
\vspace{-0.3cm}
\begin{figure}[!t]
\centering
\includegraphics[width=\linewidth, height=4.5cm]{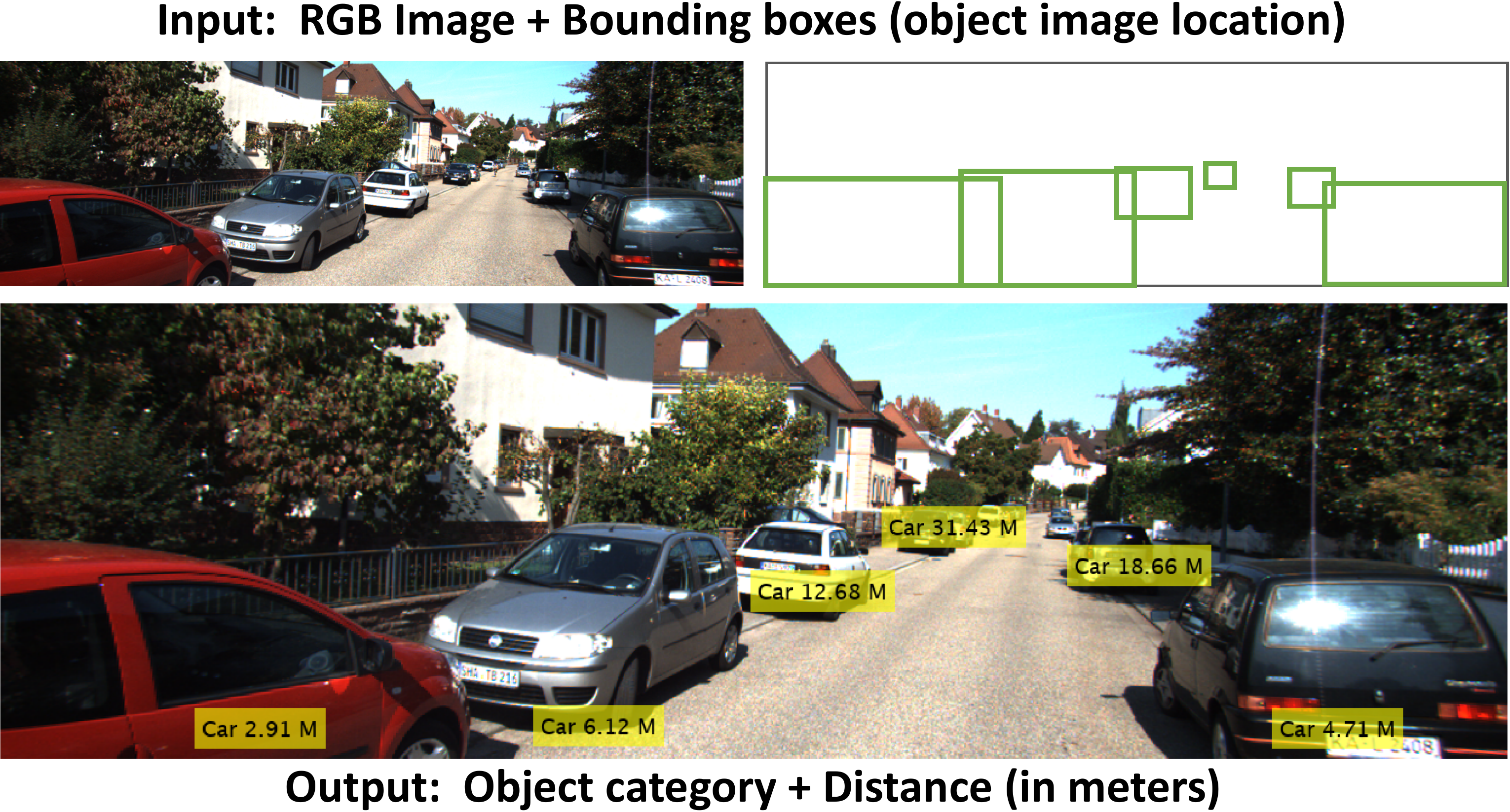}
\vspace{-0.3cm}
\caption{Given a RGB image and the bounding boxes (image location) for objects as inputs, our model directly predicts a distance (in meters) and a category label for each object in the image. Our model can be easily generalized on any visual environment reception system by appending to mature 2D detectors. }
\label{introduction}
\vspace{-0.3cm}
\end{figure}
With the advances in the field of computer vision, visual environment perception, which includes object classification, detection, segmentation and distance estimation, has become a key component in the development of autonomous driving cars. Although researchers have paid a lot of efforts on improving the accuracy of visual perception, they mainly focus on more popular tasks, such as object classification, detection and segmentation \cite{yoshioka2017real, teichman2011practical, kong2018recurrent}. Besides recognizing the objects on the road, it is also important to estimate the distances between camera sensors and the recognized objects (e.g. cars, pedestrians, cyclists), which can provide crucial information for cars to avoid collisions, adjust its speed for safety driving and more importantly, as hints for sensor fusion and path planning. However, the object-specific distance estimation task attracts very few attentions from the computer vision community. With the emergence of the convolutional neural networks, researchers have achieved remarkable progress on traditional 2D computer vision tasks using deep learning techniques, such as object detection, semantic segmentation, instance segmentation, scene reconstruction \cite{chen2018masklab, zeng20173dmatch, zhao2018icnet, hu2018relation}, but we have failed to find any deep learning application on object-specific distance estimation. One of the main reasons could be the lack of datasets that provides distance for each of the object in the images captured from the outdoor road scene. 

In this paper, we focus on addressing the interesting but challenging object-specific distance estimation problem for autonomous driving (as shown in Fig. \ref{introduction}).  We have observed that most of the current existing robotic systems or self-driving systems predict object distance by employing the traditional inverse perspective mapping algorithm. They first locate a point on the object in the image, then project the located point (usually on the lower edge of the bounding box) into a bird's-eye view coordinate using camera parameters, and finally estimate the object distance from the bird's-eye view coordinate. Though this simple method can predict reasonable distances for objects that stay close and strictly in front of the camera, it performs poorly on cases that 1) objects are located on the sides of the camera or the curved road, and 2) objects are far away (above $40$ meters) from the camera. Therefore, we are seeking to develop a model to address the aforementioned challenging cases with the advantages of deep learning techniques. 

Ours is the first work to develop an end-to-end learning-based approach that directly predicts distances for given objects in the RGB images. We build a base model that extracts features from RGB images, then utilizes ROI pooling to generate a fixed-size feature vector for each object, and finally feeds the ROI features into a distance regressor to predict a distance for each object. Though our base model is able to provide promising prediction, it still does not fulfill the precision requirement for autonomous driving. Therefore, we create an enhanced model for more precise distance estimation, particularly for objects close to the camera. Specially, in the enhanced model, we design a keypoint regressor to predict part of the 3D keypoint coordinates ($X, Y$). Together with the predicted distance ($Z$), it forms a complete 3D keypoint  ($X, Y, Z$). Leveraging the camera projection matrix, we define a projection loss between the projected 3D point and the ground truth keypoint on image to enforce a correct prediction. Note that the keypoint regressor and projection loss are used for training only. After training, given an image with object (bounding box), the object-specific distance can be directly extracted from the outputs of our trained model. There is no camera parameters intervention during inference.

To validate our proposed methods, we construct an extended dataset based on the public available KITTI object detection dataset \cite{geiger2012we}  and the newly released nuScenes (mini) dataset \cite{nuscenes2019} by computing the distance for each object using its corresponding LiDAR point cloud and camera parameters. In order to quantitatively measure the performance of our work and alternative approaches, we employ the evaluation metrics from depth prediction task as our measurements. We report the quantitative results, and visualize some examples for qualitative comparison. The experimental results on our constructed object-specific distance dataset demonstrate that our deep-learning-based models can successfully predict distances for given objects with superior performance over alternative approaches, such as the traditional inverse perspective mapping algorithm and the support vector regressor. Furthermore, our enhanced model can predict a more precise distance than our base one for objects close to the camera. The inference runtime of our proposed model is twice as fast as the traditional IPM.

In summary, the main contributions of our work are concluded as:
\renewcommand{\labelitemi}{$\bullet$}
\begin{itemize}
\item To address the object-specific distance estimation challenges, e.g., objects far away from the camera or on the curved road, we propose the first deep-learning-based method with a novel end-to-end framework (as our base model) to directly predict distance from given objects on RGB images without any camera parameters intervention. 
\vspace{-0.1cm}
\item We further design an enhanced method with a keypoint regressor, where a projection loss is introduced to improve the object-specific distance estimation, especially for object close to the camera. 
\vspace{-0.1cm}
\item To facilitate the training and evaluation on this task, we construct the extended KITTI and nuScenes (mini) object-specific distance datasets. The experiment results demonstrate that our proposed
method achieves superior performance over alternative approaches.
\end{itemize}

\begin{figure*}[!htbp]
        \centering
        \includegraphics[width=\linewidth, height=5.5cm]{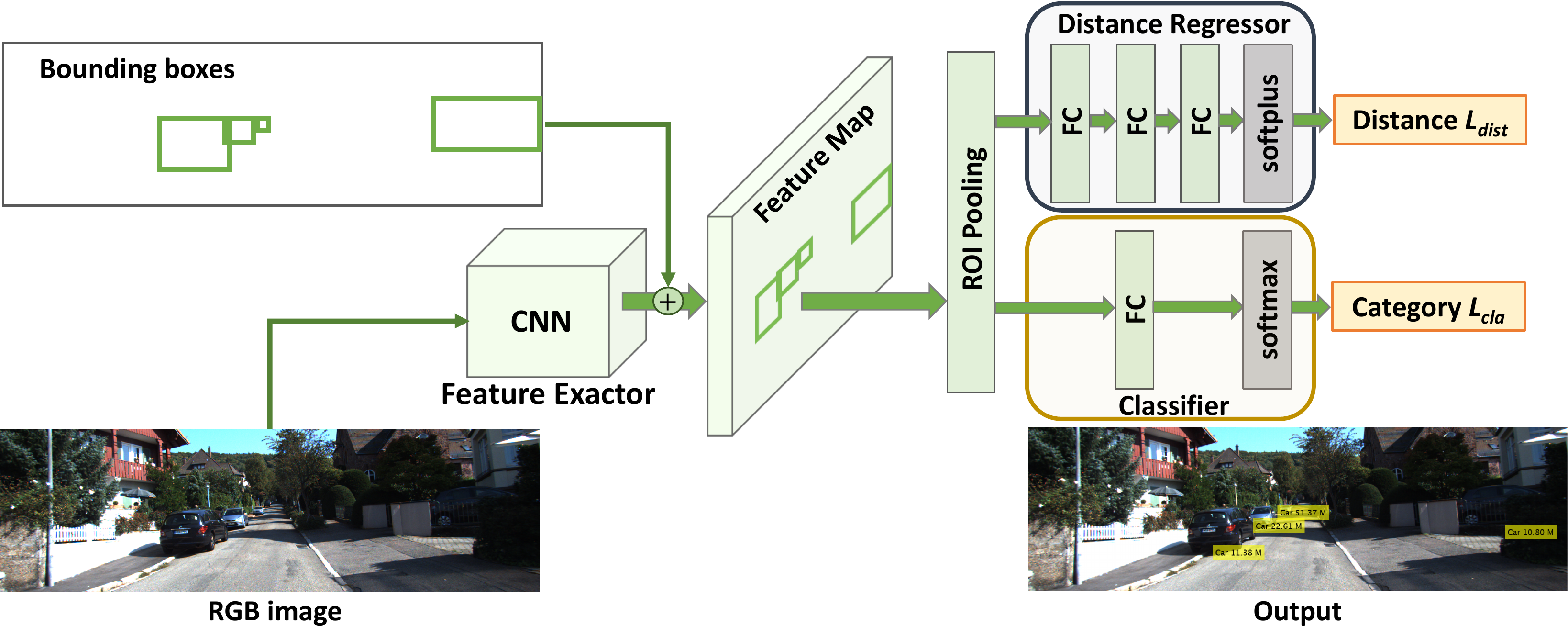}
        \vspace{-0.4cm}
        \caption{The framework of our base model, which consists of three components: a feature extractor to generate a feature map for the whole image, a distance regressor to directly predict a distance from the object specific ROI feature, and a multiclass classifier to predict the category from the ROI feature. }
        \label{baseline}
        \vspace{-0.2cm}
\end{figure*} 
\section{Related work}
Object-specific distance estimation plays a very important role in the visual environment reception for autonomous driving. In this section, we briefly review some classic methods on distance estimation and the advances of deep learning models in 2D visual perception. 

\vspace{0.1cm}
\textbf{Distance estimation} \hspace{0.3cm} Many prior works for distance estimation mainly focused on building a model to represent the geometry relation between points on images and their corresponding physical distances on the real-world coordinate. One of the classic ways to estimate distance for given object (with a point or a bounding box in the image) was to convert the image point to the corresponding bird's-eye view coordinate using inverse perspective mapping (IPM) algorithm \cite{tuohy2010distance, rezaei2015robust}. Due to the drawbacks of IPM, it would fail in cases that objects are located over 40 meters apart or on a curved road.  Another vision-based distance estimation work \cite{gokcce2015vision} learned a support vector machine regressor to predict an object-specific distance given the width and height of a bounding box. DistNet \cite{haseeb2018disnet} was a recent try to build a network for distance estimation, where the authors utilized a CNN-based model (YOLO) for bounding boxes prediction instead of the image features learning for distance estimation. Similar to IPM, their distance regressor solely studied the geometric relation that maps a bounding box with a certain width and height to a distance value. In contrast, our goal is to build a model that directly predicts distances from the learned image features. 

Besides the aforementioned approaches, some other works attempted to address this challenging problem by making use of some auxiliary information. Some marker-based methods \cite{cao2013circle, roberts2015accurate} first segmented markers in the image then estimated distance using the marker area and camera parameters.  Instead of utilizing markers, Feng et al. \cite{feng2016new} proposed a model to predict physical distance based on a rectangular pattern, where four image points of a rectangular were needed to compute the camera calibration. They then predicted the distance of any given point on an object using the computed camera calibration. Though prior works are impressive, they require markers or patterns to be put in the image for distance estimation, which limits their generalization for autonomous driving. 
\begin{figure*}[htbp]
        \centering
        \vspace{-0.3cm}
        \includegraphics[width=\linewidth, height=8.5cm]{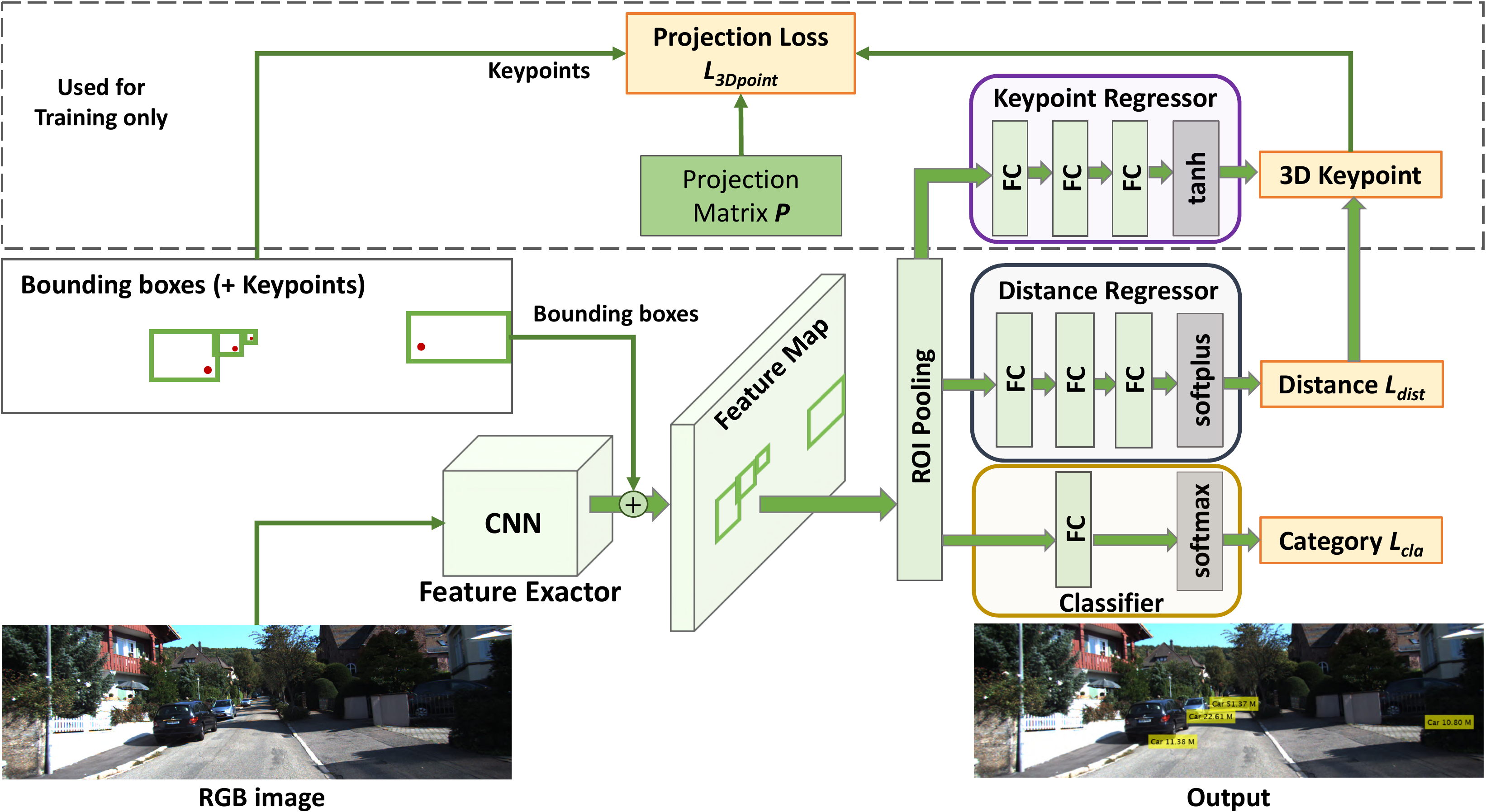}
        \caption{The framework of our enhanced model, which contains four parts, a feature extractor to generate a feature map for the whole RGB image,  a keypoint regressor to predict a keypoint position on 3D coordinate, a distance regressor  to directly predict a distance , and a multiclass classifier to predict the category label. The outputs of the keypoint regressor and distance regressor compose a 3D keypoint, which will be projected back to the image plane using the camera projection matrix. A projection loss is defined between the projected keypoint and the ground truth keypoint to enforce a better distance estimation.}
        \label{pipeline}
         \vspace{-0.2cm}
\end{figure*} 

\vspace{0.1cm}
\textbf{2D visual perception}  \hspace{0.3cm}  Although there is no recent work employing deep learning techniques to learn the robust image features for visual monocular object-specific distance estimation, deep learning techniques have been successfully applied on many other 2D visual perception tasks (e.g. object detection, classification, segmentation, monocular depth estimation) with excellent performance \cite{AAAIZhu, Chen_2019_CVPR, dai2017deep, zhu2015learning}. The series of R-CNN works \cite{girshick2014rich, girshick2015fast, ren2015faster, he2017mask} are the pioneers to boost the accuracy as well as decrease processing time consumption for object detection, classification and segmentation. SSD \cite{liu2016ssd} and YOLO models \cite{redmon2016you, redmon2017yolo9000} are also the popular end-to-end frameworks to detect and classify objects in RGB images. Their models could be used to address some of the visual perception tasks for autonomous driving, such as detection and classification, but their models are unable to predict the object distance. Nevertheless, those remarkable works inspired us to build an effective end-to-end model for monocular object-specific distance estimation.  

On the other hand, monocular depth estimation could be a problem close to our object-specific distance estimation task. Recently, many researchers have created some supervised and even unsupervised models to predict dense depth maps for given monocular color images with more precise details \cite{eigen2014depth, liu2015deep, liu2016learning, garg2016unsupervised}. Their works are motivating, but they usually cost more memory and processing time no matter if it is for training or testing. For visual perception of autonomous driving, it is more crucial to know the object-specific distance to avoid collisions or fuse multiple sensor information, instead of the dense depth map for the entire scene. 

\section{Our Approach} \label{Our Approach}
Observing the limits of the classic inverse mapping algorithm on distance estimation, we propose a learning-based model for robust object-specific distance estimation. A model that directly predicts the physical distance from given RGB images and object bounding boxes,  is introduced as our base model. Moreover, we design an enhanced model with a keypoint regressor for a better object-specific distance estimation.  

\subsection{Base method} \label{Baseline method}
Our base model consists of three components, i.e., a feature extractor, a distance regressor and a multiclass classifier (as shown in Fig. \ref{baseline}). 

\vspace{0.1cm}
\textbf{Feature extractor} \hspace{0.3cm} In our model, a RGB image is fed into an image feature learning network to extract the feature map for the entire RGB image. We exploit the popular network structures (e.g., vgg16, res50) as our feature extractor. The output of the last layer of CNN will be max-pooled and then extracted as the feature map for the given RGB image. 

\vspace{0.1cm}
\textbf{Distance regressor and classifier}  \hspace{0.3cm}  
We feed the extracted feature map from feature extractor and the object bounding boxes (implying the object locations in the image) into an ROI pooling layer to generate a fixed-size feature vector $\bm{F_i}$ to represent each object in the image. The pooled feature then is passed through the distance regressor and classifier to predict a distance and a category label for each object. The distance regressor contains three fully connected (FC) layers (with layers of size $\{2048,512,1\}$ for vgg16, $\{1024,512,1\}$ for res50). A softplus activation function is applied on the output of the last fully connected layer to make sure the predicted distance (denoted as $D(\bm{F_i})$ is positive. For the classifier, there is a fully connected (FC) layer (with the neuron size equals to the number of the categories in the dataset) followed by a softmax function. Let the output of the classifier be $C(\bm{F_i})$. Our loss for the distance regressor $\mathcal{L}_{dist}$ and classifier $\mathcal{L}_{cla}$ can be written as:   
\vspace{-0.3cm}
\begin{equation}
\vspace{-0.2cm}
\label{eq1}
\begin{split}
&\mathcal{L}_{dist} =  \frac{1}{N}\sum_{i=1}^N \text{smooth}_{L1}(d_{i}^* - D(\bm{F_i})), \\
\end{split}
\end{equation}
\begin{equation}
\label{eq2}
\begin{split}
&\mathcal{L}_{cla} = \frac{1}{N}\sum_{i=1}^N \text{cross-entropy} (y_i^*, C(\bm{F_i})), \\
\end{split}
\end{equation}
where $N$ is the number of objects, $d_{i}^*$ and $y_i^*$ are the ground truth distance and category label for the $i$-th object . 

\vspace{0.1cm}
\textbf{Model learning and inference} \hspace{0.3cm} We train the feature extractor, the distance regressor and the classifier simultaneously with loss 
\vspace{-0.3cm}
\begin{equation}
\label{eq3}
\begin{split}
\min \mathcal{L}_{base} =  \mathcal{L}_{cla} + \lambda_1 \mathcal{L}_{dist}. 
\end{split}
\end{equation}
We use ADAM optimizer to obtain the optimal network parameters with beta value $\beta = 0.5$. The learning rate is initialized as $0.001$ and exponentially decayed after $10$ epochs.  $\lambda_1$ is set to $1.0$  when training our framework. Note that the classifier network is used during training only. Implying a prior knowledge of the correlation between the object class and its real size and shape, the classifier encourages our model to learn features that can be leveraged in estimating more accurate distances. After training, our base model can be used to directly predict the object-specific distances given any RGB images and object bounding boxes as input.

\subsection{Enhanced method} \label{Enhanced method}
Though our base model is able to predict promising object-specific distance from ROI feature map, it is still not satisfying the precision requirement for autonomous driving, especially for objects close to the camera. Therefore, we design an enhanced method with a keypoint regressor to optimize the base model by introducing a projection constraint, and as a result to enforce a better distance prediction.  As shown in Fig. \ref{pipeline}, the pipeline of our enhanced model consists of four parts, a feature extractor, a keypoint regressor, a distance regressor and a multiclass classifier. 

\vspace{0.1cm}
\textbf{Feature extractor} \hspace{0.3cm} We utilize the same network structure that we use in our base model to extract the RGB image feature. With the object bounding boxes, we can obtain the object-specific features $\bm{F_i}$ using ROI-pooling (see Sec. \ref{Baseline method} for details).

\vspace{0.1cm}
\textbf{Keypoint regressor} \hspace{0.3cm} The keypoint regressor $K$ learns to predict an approximate keypoint position in the 3D camera coordinate system. The output of the distance regressor can be considered as the value on the camera $\emph{Z}$ coordinate, so there are only two coordinate values ($\emph{X, Y}$) that need to be predicted by the keypoint regressor, denoted as $K(\bm{F_i})$. It contains three fully connected (FC) layers of sizes $\{2048, 512, 2\}$, $\{1024, 512, 2\}$ for vgg16 and res50, respectively. Since we do not have the ground truth of the 3D keypoint, we choose to project the generated 3D point ([$K(\bm{F_i}), D(\bm{F_i})$]) back to the image plane using the camera projection matrix $P$. Then we compute the errors between the ground truth 2D keypoint $k_i^*$ and the projected point ($P\cdot[K(\bm{F_i}), D(\bm{F_i})]$). In order to encourage the model to better predict distances for closer objects, we put a weight with regard to the ground truth distance into the projection loss $\mathcal{L}_{3Dpoint}$ as 
\vspace{-0.3cm}
\begin{equation}
\label{eq4}
\begin{split}
&\mathcal{L}_{3Dpoint} = \frac{1}{N}\sum_{i=1}^N \frac{1}{d_{i}^*} ||P\cdot [K(\bm{F_i}), D(\bm{F_i})] - k_i^*||_2. \\
\end{split}
\end{equation} 

\vspace{0.1cm}
\textbf{Distance regressor and classifier} \hspace{0.3cm}  For the distance regressor and classifier , we leverage the same network structure as well as training loss  $\mathcal{L}_{dist}$ (Eq. \ref{eq1}) and $\mathcal{L}_{cla}$ (Eq. \ref{eq2}) as the base model. The network parameters in the distance regressor are optimized by the projection loss $\mathcal{L}_{3Dpoint}$ as well. 

\vspace{0.1cm}
\textbf{Network learning and inference} \hspace{0.3cm} We train the feature extractor, the keypoint regressor, the distance regressor and the classifier simultaneously with loss 
\vspace{-0.3cm}
\begin{equation}
\vspace{-0.1cm}
\label{eq5}
\begin{split}
\min \mathcal{L}_{enhance} =  &\mathcal{L}_{cla} + \lambda_1 \mathcal{L}_{dist} + \lambda_2 \mathcal{L}_{3Dpoint}. \\
\end{split}
\end{equation}

We use the same setting for the optimizer, beta value and learning rate as the base model. 
$\lambda_1$,  $\lambda_2$ are set to $10.0$, $0.05$. We only use the camera projection matrix $P$, keypoint regressor and classifier for training. When testing, given a RGB image and the bounding boxes, our learned enhanced model directly predicts the object-specific distances without any camera parameters intervention. We implement our (base and enhanced) models using the popular deep learning platform PyTorch \cite{paszke2017automatic} and run them on a machine with Intel Xeon E5-2603 CPU and NVIDIA Tesla K80 GPU.

\begin{figure*}
\begin{subfigure}{.66\textwidth}
  \centering
  \vspace{-0.3cm}
  \includegraphics[width=.9\linewidth, height=6.5cm]{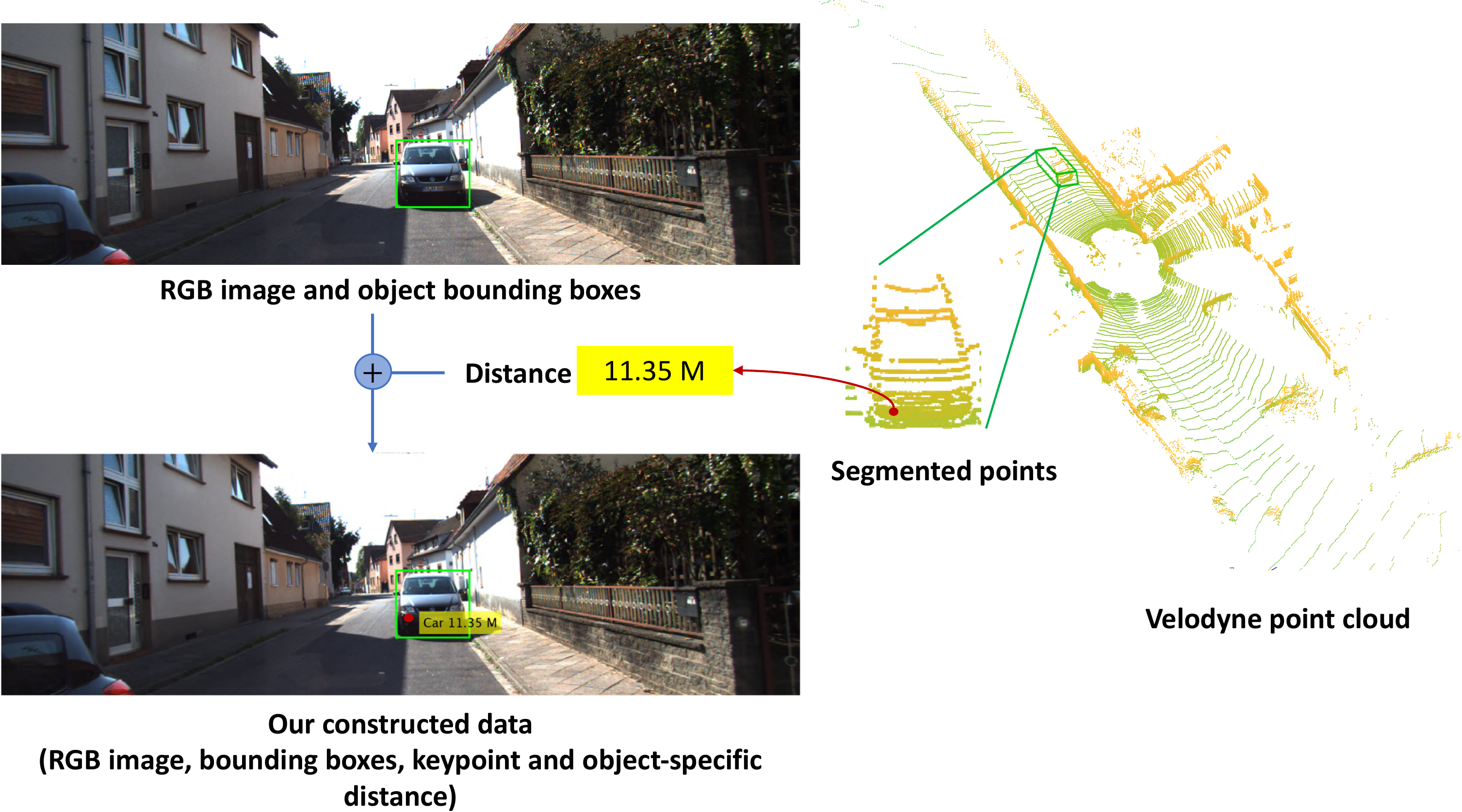}
  \caption{The pipeline of our dataset construction. For each object in the RGB image, we segment its 3D points from the corresponding velodyne point cloud and extract the depth value of the $n$-th point as the ground truth distance. We project the $n$-th point to the image plane to get the 2D keypoint coordinates. Both the extracted distance and the 2D keypoint coordinate of the $n$-th velodyne point are added into the KITTI / nuScenes(mini) object detection dataset as the extension.}
  \label{datasetConstruction}
\end{subfigure}
\hspace{0.3cm}
\begin{subfigure}{.32\textwidth}
  \centering
  \includegraphics[width=\linewidth,  height=5.8cm]{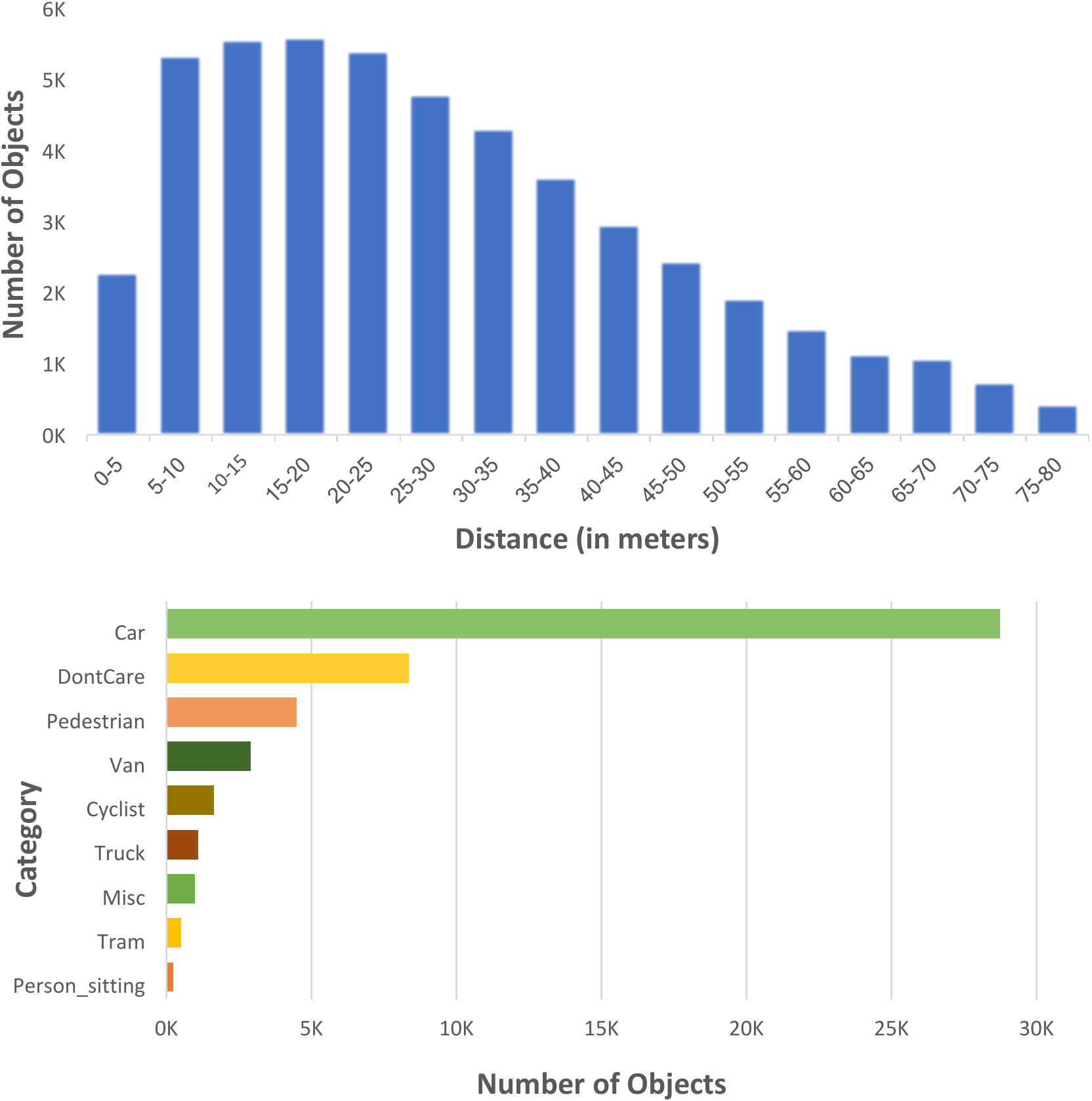}
  \caption{Distribution of generated object-specific distances (top), and different object categories in our constructed KITTI-based object-specific distance dataset (bottom).}
  \label{datasetAnalysis}
\end{subfigure}
\vspace{-0.3cm}
\caption{Our dataset construction strategy and the distributions. Fig. \ref{datasetConstruction} is the pipeline how we construct our dataset with generated ground truth object-specific distances, while Fig. \ref{datasetAnalysis} shows the distribution of the generated KITTI-based object-specific distances and the object categories.}
\label{pipeConstruction}
 \vspace{-0.2cm}
\end{figure*}

\section{Training data construction} \label{Training data construction}
One of the main challenges of training deep neural networks for object-specific distance estimation task is the lack of datasets with distance annotation for each object in the RGB images.  Existing object detection datasets only provide the bounding boxes and object category annotations, while dense depth prediction datasets provide pixel-level depth values for each image without any object information. Neither of them provide clear object-specific distance annotations. Therefore, we construct two extended object detection datasets from KITTI and nuScenes (mini) with ground truth object-specific distance for autonomous driving. 

\vspace{0.1cm}
\textbf{KITTI and nuScenes (mini) dataset} \hspace{0.3cm} As one of the well-known benchmark datasets for autonomous driving, KITTI \cite{geiger2012we} provides an organized dataset for object detection task with RGB image, bounding (2D and 3D) boxes, category labels for objects in the images, and the corresponding velodyne point cloud for each image, which is ideal for us to construct a object-specific distance dataset. Similarly, the newly released nuScenes(mini) \cite{nuscenes2019} also contains all the information (i.e., RGB images, bounding boxes, velodyne point clouds) for our dataset construction. 

\vspace{0.1cm}
\textbf{Object distance ground truth generation} \hspace{0.3cm} As shown in Fig. \ref{datasetConstruction}, to generate the object-specific distance ground truth for a object in a RGB image, we first segment the object points from the corresponding velodyne point cloud using its 3D bounding box parameters; then sort all the segmented points based on their depth values; and finally exact the $n$-th depth value from the sorted list as the ground truth distance for given object. In our case,  we set $n=0.1\times(\text{number of segmented points})$ to avoid extracting depth values from noise points. Additionally, we project the velodyne points (used for ground truth distance extraction) to their corresponding RGB image planes, and get their image coordinates as the keypoint ground truth. We append the ground truth of the object-specific distance and keypoint to the KITTI / nuScenes(mini) object detection dataset labels, together with the RGB images to construct our dataset. 

Since both KITTI and nuScenes(mini) only provide the ground truth labels for the training set in its object detection dataset, we generate the distance and keypoint ground truth for all the samples in the training set. Following the split strategy as \cite{chen2017multi}, we split the samples from KITTI training set into two subsets (training / validation) with $1:1$ ratio. There is a total of $3, 712$ RGB images with  $23, 841$ objects in the training subset, and $3, 768$ RGB images with $25, 052$ objects in the validation subset. All the objects are categorized into $9$ classes, i.e., \emph{Car, Cyclist, Pedestrian, Misc, Person\_sitting, Tram, Truck, Van, DontCare}. Our generated ground truth object-specific distances are varied from $[0, 80]$ in meters. Fig. \ref{datasetAnalysis} shows the distribution of the generated object-specific distances and the object categories in our entire constructed dataset. We can find that distances are ranged mostly from $5$M to $60$M, and \emph{Car} is the dominant category in the dataset. For the nuScenes(mini) dataset, we randomly split the samples into two subsets with $1,549$ objects in $200$ training images and $1,457$ objects in $199$ validation images. All objects are labeled with $8$ categories (\emph{Car, Bicycle, Pedestrian, Motorcycle, Bus, Trailer, Truck, Construction\_vehicle}) and distances varied from $2$M to $105$M.

\section{Evaluation}
In this section, we evaluate our proposed models with a comparison to alternative approaches.  We train our models on the training subsets of our constructed datasets, while test them on the validation subsets.

\vspace{0.1cm}
\textbf{Evaluation metrics} \hspace{0.3cm} Our goal is to predict a distance for objects as close to the ground truth distance as possible. Therefore, we adopt the evaluation metrics provided by \cite{eigen2014depth}, usually used for depth prediction. It includes absolute relative difference (\emph{Abs Rel}),  squared relative difference (\emph{Squa Rel}), root of mean squared errors (\emph{RMSE}) and root of mean squared errors computed from the log of the predicted distance and the log ground truth distance (\emph{RMSE$_{log}$}). Let $d_i^{*}$ and $d_i$ denote the ground truth distance and the predicted distance, we can compute the errors as 
\vspace{-0.2cm}
\begin{align}
    &\text{Threshold: $\%$ of }d_i\,\, s.t. \max({d_i}/{d_i^{*}}, {d_i^{*}}/{d_i})=\delta< \text{\emph{threshold}} ,\notag\\
    \vspace{-0.1cm}
    &\text{Abs Relative difference (\emph{Abs Rel}): }\frac{1}{N}\sum_{d\in N}{|d-d^{*}|}/{d^{*}},\notag\\
    \vspace{-0.1cm}
    &\text{Squared Relative difference (\emph{Squa Rel}): } \frac{1}{N}\sum_{d\in N}{||d-d^{*}||^2}/{d^{*}},\notag\\
    \vspace{-0.1cm}
    &\text{RMSE (linear)}: \sqrt{\frac{1}{N}\sum_{d\in N}||d_i - d_i^{*}||^2},\notag\\   
    \vspace{-0.1cm}
    &\text{RMSE (log)}: \sqrt{\frac{1}{N}\sum_{d\in N}||\log d_i - \log d_i^{*}||^2}.\notag
\end{align}

\vspace{0.1cm}
\textbf{Compared approaches} \hspace{0.3cm}  
As one of the most classic methods to predict (vehicle) distance in an automobile environment, inverse perspective mapping algorithm (IPM) \cite{tuohy2010distance} approximates a transformation matrix between a normal RGB image and its bird's-eye view image using camera parameters. We adopt the IPM in the MATLAB computer vision toolkit to get the transformation matrices for the RGB images (from validation subset). After projecting the middle points of the lower edge of the object bounding boxes into their bird's-eye view coordinates using the IPM transformation matrices, we take the values along forward direction as the estimated distances. 

Similar to the recent work \cite{gokcce2015vision}, we compute the width and height of each bounding box in the training subset, and train a SVR with the ground truth distance. After that, we get the estimated distances for objects in the validation set by feeding the widths and heights of their bounding boxes into the trained SVR.   

For our proposed model, we utilize vgg16 and res50 as our feature extractor for both base and enhanced model. We trained our models for $20$ epochs with the batch size of $1$ on the training dataset augmented with horizontally-flipped training images.  After training, we feed the RGB image with the bounding boxes into our trained models and take the output of the distance regressor as the estimated distance for each object in the validation subset. 
\begin{figure}
\centering
\vspace{-0.1cm}
\includegraphics[width=\linewidth, height=2.1cm]{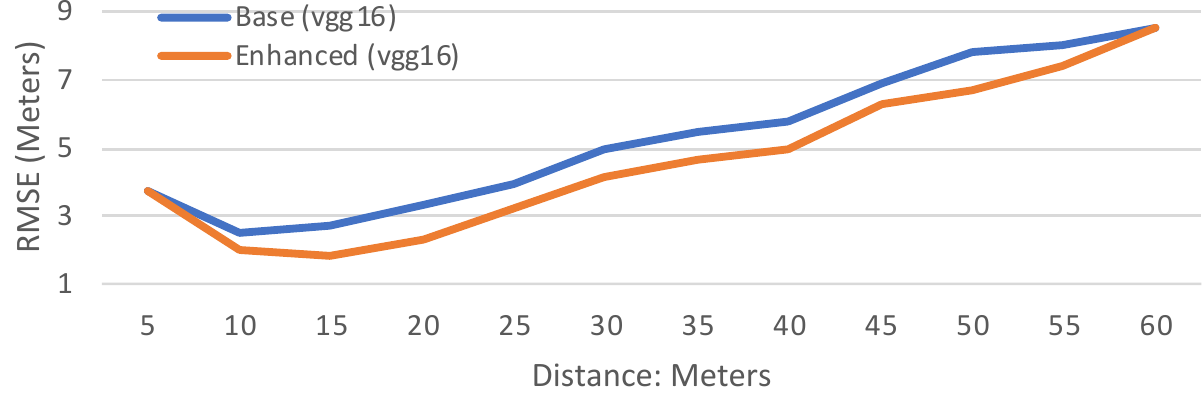}
\vspace{-0.6cm}
\caption{Average RMSE on objects with different distances in the KITTI-based  dataset (lower is better).}
\label{compare}
 \vspace{-0.5cm}
\end{figure}
\begin{table*}[!ht]	
	\footnotesize
	\centering
	\vspace{-0.3cm}
	\caption{The comparisons of object-specific distance estimation with alternative approaches on the \emph{val} subset of our constructed KITTI-object-detection-based dataset.}
	\vspace{-0.3cm}
	\label{AllResult_Table}
	\begin{tabular}{cccccccccc}
		\hline
		\hline
		\noalign{\smallskip}
		&\multirow{2}{*}{\textbf{Method}} &\multicolumn{3}{c}{higher is better}&&\multicolumn{4}{c}{lower is better} \\
		\cline{3-5}\cline{7-10}
		\noalign{\smallskip}
		&&$\delta < 1.25$&$\delta < 1.25^2$&$\delta < 1.25^3$&&Abs Rel&Squa Rel&RMSE&RMSE$_{log}$\\
		\hline\noalign{\smallskip}
		\multirow{7}{*}{\textbf{Car}}&\multicolumn{1}{l}{Support Vector Regressor (SVR) \cite{gokcce2015vision}} &0.345 &0.595 &0.823 &&1.494 &47.748 &18.970 &1.494 \\	
		&\multicolumn{1}{l}{Inverse Perspective Mapping (IPM) \cite{tuohy2010distance}} &0.701 &0.898 &0.954 &&0.497 &1290.509 &237.618 &0.451  \\				
		\cline{2-10}\noalign{\smallskip}
		&\multicolumn{1}{l}{\textbf{Our Base Model (res50)}} &0.782 &0.927 &0.964 &&0.178 &0.843 &4.501 &0.415  \\
		&\multicolumn{1}{l}{\textbf{Our Base Model (vgg16)}} &0.846 &\textbf{0.947} &\textbf{0.981} &&\textbf{0.150} &\textbf{0.618} &3.946 &\textbf{0.204}  \\
		&\multicolumn{1}{l}{\textbf{Our Enhanced Model (res50)}} &0.796 &0.924 &0.958 &&0.188 &0.843 &4.134 &0.256 \\
		&\multicolumn{1}{l}{\textbf{Our Enhanced Model (vgg16)}} &\textbf{0.848} &0.934 &0.962 &&0.161 &0.619 &\textbf{3.580} &0.228 \\		
		\hline\noalign{\smallskip}
		\multirow{7}{*}{\textbf{Pedestrian}}&\multicolumn{1}{l}{Support Vector Regressor (SVR) \cite{gokcce2015vision}} &0.129 &0.182 &0.285 &&1.499 &34.561 &21.677 &1.260 \\	
		&\multicolumn{1}{l}{Inverse Perspective Mapping (IPM) \cite{tuohy2010distance}} &0.688 &0.907 &0.957 &&0.340 &543.223 &192.177 &0.348 \\			
		\cline{2-10}\noalign{\smallskip}
		&\multicolumn{1}{l}{\textbf{Our Base Model (res50)}} &0.649 &0.896 &0.966 &&0.247 &1.315 &4.166 &0.335   \\
		&\multicolumn{1}{l}{\textbf{Our Base Model (vgg16)}} &0.578 &0.861 &0.960 &&0.289 &1.517 &4.724 &0.312  \\
		&\multicolumn{1}{l}{\textbf{Our Enhanced Model (res50)}} &0.734 &\textbf{0.963} &\textbf{0.988} &&0.188 &0.807 &3.806 &0.225 \\
		&\multicolumn{1}{l}{\textbf{Our Enhanced Model (vgg16)}} &\textbf{0.747} &0.958 &0.987 &&\textbf{0.183} &\textbf{0.654} &\textbf{3.439} &\textbf{0.221} \\
		\hline\noalign{\smallskip}	
		\multirow{7}{*}{\textbf{Cyclist}}&\multicolumn{1}{l}{Support Vector Regressor (SVR) \cite{gokcce2015vision}} &0.226 &0.393 &0.701 &&1.251 &31.605 &20.544 &1.206  \\	
		&\multicolumn{1}{l}{Inverse Perspective Mapping (IPM) \cite{tuohy2010distance}} &0.655 &0.796 &0.915 &&0.322 &9.543 &19.149 &0.370 \\				
		\cline{2-10}\noalign{\smallskip}
		&\multicolumn{1}{l}{\textbf{Our Base Model (res50)}} &0.744 &0.938 &0.976 &&0.196 &1.097 &4.997 &0.309   \\
		&\multicolumn{1}{l}{\textbf{Our Base Model (vgg16)}} &0.740 &0.942 &0.979 &&0.193 &0.912 &\textbf{4.515} &0.240  \\
		&\multicolumn{1}{l}{\textbf{Our Enhanced Model (res50)}} &0.766 &0.947 &\textbf{0.981} &&\textbf{0.173} &\textbf{0.888} &4.830 &\textbf{0.225} \\
		&\multicolumn{1}{l}{\textbf{Our Enhanced Model (vgg16)}} &\textbf{0.768} &\textbf{0.947} &0.974 &&0.188 &0.929 &4.891 &0.233 \\
		\hline\noalign{\smallskip}
		\multirow{7}{*}{\textbf{Average}}&\multicolumn{1}{l}{Support Vector Regressor (SVR) \cite{gokcce2015vision}} &0.379 &0.566 &0.676 &&1.472 &90.143 &24.249 &1.472 \\	
		&\multicolumn{1}{l}{Inverse Perspective Mapping (IPM) \cite{tuohy2010distance}} &0.603 &0.837 &0.935 &&0.390 &274.785 &78.870 &0.403  \\			
		\cline{2-10}\noalign{\smallskip}
		&\multicolumn{1}{l}{\textbf{Our Base Model (res50)}} &0.503 &0.776 &0.905 &&0.335 &3.095 &8.759 &0.502   \\
		&\multicolumn{1}{l}{\textbf{Our Base Model (vgg16)}} &0.587 &0.812 &0.918 &&0.311 &2.358 &7.280 &0.351   \\
		&\multicolumn{1}{l}{\textbf{Our Enhanced Model (res50)}} &0.550 &0.834 &\textbf{0.937} &&0.271 &2.363 &8.166 &0.336 \\
		&\multicolumn{1}{l}{\textbf{Our Enhanced Model (vgg16)}} &\textbf{0.629} &\textbf{0.856} &0.933 &&\textbf{0.251} &\textbf{1.844} &\textbf{6.870} &\textbf{0.314} \\
		\hline
		\hline
	\end{tabular}
	 \vspace{-0.3cm}
\end{table*}

\vspace{0.1cm}
\textbf{Results on KITTI dataset}  \hspace{0.3cm} We present a quantitative comparison in the constructed KITTI dataset for all the evaluation metrics in Table \ref{AllResult_Table}. Note that we do not include the distances predicted for \emph{DontCare} objects when calculating the errors. In addition to the average errors among the $8$-category objects, we also provide the performance on three particular categories, i.e., \emph{Car}, \emph{Pedestrian}, \emph{Cyclist}, for comprehensive analysis. As we can see from the table, our proposed models are able to predict distances with much lower relative errors and higher accuracy when compared with the IPM and SVR. Moreover, our enhanced model performs the best among all the compared methods, which implies the effectiveness of the introduction of keypoint regressor and projection constraint. Besides, our models perform pretty well on \emph{Car}, \emph{Pedestrian}, \emph{Cyclist} objects but with a slightly worse average performance. We have investigated the results on each category, and found that our models perform relatively poor on some categories with fewer training samples, such as \emph{Person\_sitting}, \emph{Tram}. 
Fig. \ref{compare} clearly illustrates the improvement of the enhanced model on objects with different distances.

\begin{figure*}[!htbp]
        \centering
       \vspace{-0.3cm}
        \includegraphics[width=\linewidth]{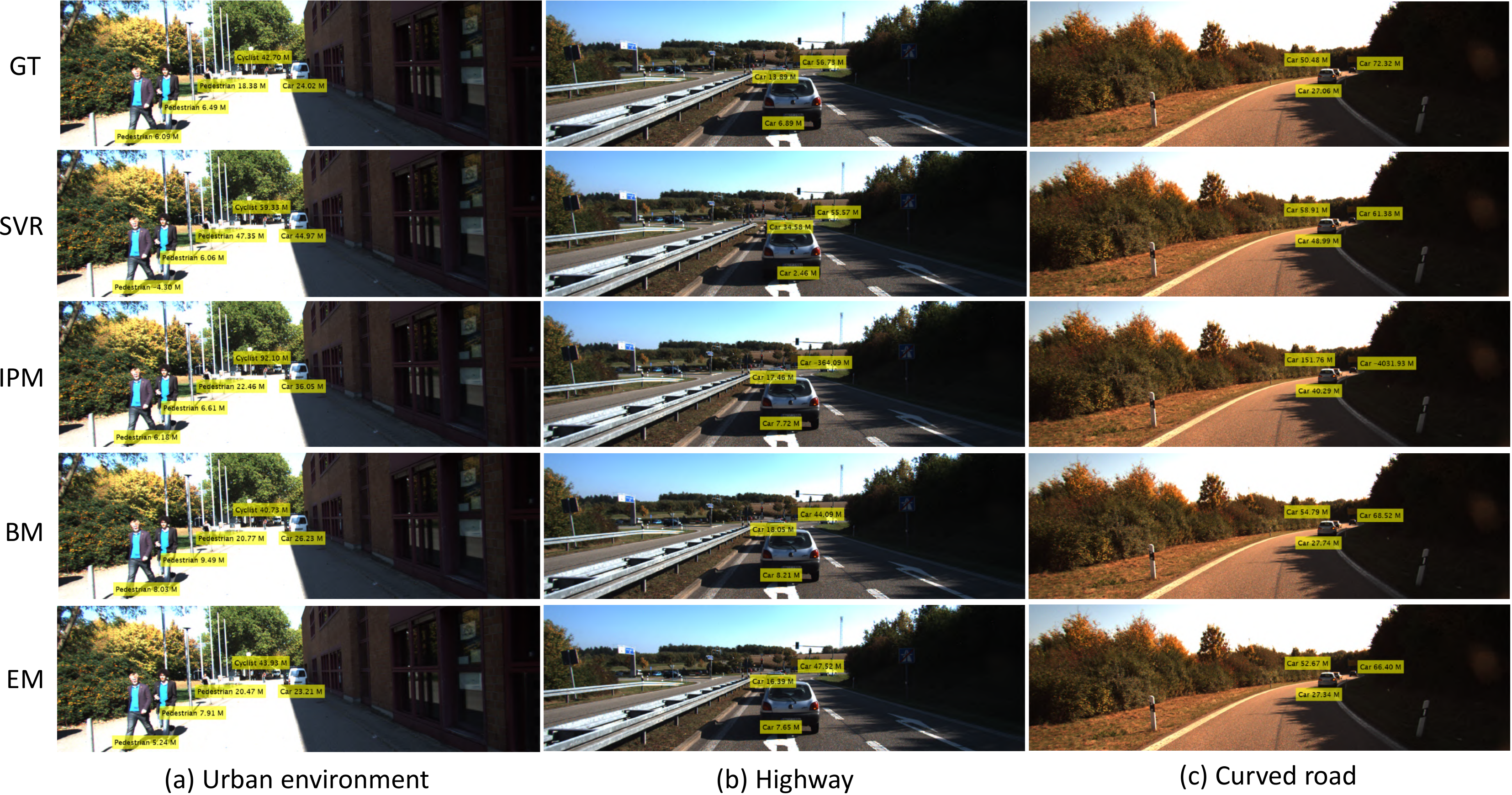}
        \caption{Examples of the estimated distance using our proposed base model (BM) and enhanced model (EM). We also provide ground truth distance (GT), the predicted distances using IPM and SVR for comparison. Our models can successfully predict distances on challenging cases, such as objects over $40$ meters or on the curved road. }
       \label{results}
        \vspace{-0.3cm}
\end{figure*}

In addition to the quantitative comparison, we visualize some estimated object-specific distance using our proposed models, along with the ground truth distance and the predictions using alternative IPM and SVR for comparison in Fig. \ref{results}. The SVR results show the difficulties to estimate a distance according to the width and height of a bounding box. IPM usually performs well for the objects close to or strictly in front of the camera, while it generally predicts incorrect distances for objects far away from the camera, such as the cyclist on the urban environment example, the furthest cars on both highway and curved road images. However, both of our models can predict more accurate distances for those objects. The other challenging case is to predict distance for objects on a curved road. IPM fails when vehicles are turning, whereas our models can successfully handle them. Besides, our enhanced model predicts a more precise objects-specific distance with less time. The average inference time of our model (vgg16) is $16.2ms$ per image, which is slightly slower than SVR ($12.1 ms$) but twice as fast as IPM ($33.9 ms$). 
\begin{table}[]	
	\scriptsize
	\centering
	\vspace{-0.1cm}
	\caption{Comparison of our models trained with and without the classifier on (average) KITTI distance estimation.}
	\vspace{-0.35cm}
	\label{Ablation_Table}
	\setlength{\tabcolsep}{1.7mm}{
	\begin{tabular}{p{0.3cm}p{0.4cm}p{0.3cm}p{0.3cm}p{0.1mm}p{0.3cm}p{0.35cm}p{0.5cm}p{0.7cm}}
		\hline
		\multirow{2}{*}{\textbf{Vgg16 models}} &\multicolumn{3}{c}{higher is better}&&\multicolumn{4}{c}{lower is better}\\
		\cline{2-4}\cline{6-9}
		&\multicolumn{1}{c}{$\delta_1$}&$\delta_2$&$\delta_3$&&AR&SR&RMSE&RMSE$_{log}$\\
		\hline
		\multicolumn{1}{l}{Base w/o classifier} &0.482 &0.692 &0.802 &&0.658 &7.900 &9.317 &0.573 \\
		\multicolumn{1}{l}{Base w classifier} &0.587 &0.812 &0.918 &&0.311 &2.358 &7.280 &0.351   \\
		\multicolumn{1}{l}{Enhanced w/o classifier} &0.486 &0.738 &0.844 &&0.541 &5.555 &8.747 &0.512\\ 
		\multicolumn{1}{l}{\textbf{Enhanced w classifier}} &\textbf{0.629} &\textbf{0.856} &\textbf{0.933} &&\textbf{0.251} &\textbf{1.844} &\textbf{6.870} &\textbf{0.314} \\
	    \hline
	\end{tabular}}
	\vspace{-0.3cm}
    \end{table}
   
The purpose of the classifier is to encourage  our model to learn the category-discriminative features that can be useful in getting a better estimate of how far the object is. We train our (vgg16) models with and without the classifier, then compute the errors for the estimated distance on samples in the validation set. The prediction results are reported in Table \ref{Ablation_Table} under the same evaluation metrics as in Table \ref{AllResult_Table}. The performance enhancement demonstrates the effectiveness of our classifier for learning a model on object-distance estimation.

\begin{table}[]	
	\scriptsize
	\centering
	\vspace{-0.1cm}
	\caption{Comparison of (average) object-specific distance estimation on the nuScenes-based (mini) dataset.}
	\vspace{-0.35cm}
	\label{nuScenes_Table}
	\setlength{\tabcolsep}{1.7mm}{
	\begin{tabular}{p{0.1cm}p{0.3cm}p{0.3cm}p{0.3cm}p{0.01mm}p{0.3cm}p{0.65cm}p{0.5cm}p{0.35cm}}
		\hline
		\multirow{2}{*}{\textbf{Methods}} &\multicolumn{3}{c}{higher is better}&&\multicolumn{4}{c}{lower is better}\\
		\cline{2-4}\cline{6-9}
		&\multicolumn{1}{c}{$\delta_1$}&$\delta_2$&$\delta_3$&&AR&SR&RMSE&RMSE$_{log}$\\
		\hline
		\multicolumn{1}{l}{SVR \cite{gokcce2015vision}} &0.308  &0.652 &0.833 &&0.504 &13.197 &18.480 &0.846   \\
		\multicolumn{1}{l}{IPM \cite{tuohy2010distance}} &0.441 &0.772 &0.875 &&1.498 &1979.375 &249.849 &0.926 \\
		\multicolumn{1}{l}{\textbf{Base Model(res50)}} &0.310  &0.621 &0.846 &&0.466 &7.593 &15.703 &0.492 \\	
		\multicolumn{1}{l}{\textbf{Base Model(vgg16)}} &0.393  &0.697 &0.914 &&0.404 &5.592 &12.762 &0.420 \\
		\multicolumn{1}{l}{\textbf{Enhanced Model(res50)}} &0.367  &0.683 &0.877 &&0.340 &5.126 &14.139 &0.433 \\
		\multicolumn{1}{l}{\textbf{Enhanced Model(vgg16)}} &\textbf{0.535} &\textbf{0.863} &\textbf{0.959} &&\textbf{0.270} &\textbf{3.046} &\textbf{10.511} &\textbf{0.313} \\
	    \hline
	\end{tabular}}
	\vspace{-0.3cm}
    \end{table}

\vspace{0.1cm}
\textbf{Results on nuScenes dataset} \hspace{0.3cm} After training our proposed models on the training subset of the constructed nuScenes(mini) dataset, we calculate the distance estimation errors and accuracies on objects in the testing subset (as reported in Table \ref{nuScenes_Table}) using the same measurements in Table \ref{AllResult_Table}. Our enhanced model achieves the best performance among all the compared methods for object-specfic distance estimation. 

\section{Conclusion}
In this paper, we discuss the significant but challenging object-specific distance estimation problem in autonomous driving. It is the first attempt to utilize deep learning techniques for object-specific distance estimation. We introduce a base model to directly predict distances (in meters) from a given RGB image and object bounding boxes. Moreover, we design an enhanced model with keypoint projection constraint for a more precise estimation, particular for the objects close to the camera. We trained our models on our newly constructed dataset extended from KITTI and nuScenes(mini) with a ground truth distance for each object in the RGB images. The experimental results demonstrate that our base model is able to predict distances with superior performance over alternative approaches IPM and SVR,  while our enhanced model obtains the best performance over all the compared methods.

\section{Acknowledgement}
The authors would like to highly acknowledge the research team from XMotors.ai for their contributions to bring the inspiration for this proposed research direction. We appreciate Dr. Husam Abu-haimed and Dr. Junli Gu for their valuable inputs to initialize this research work. We also particularly thank Dr. Husam Abu-haimed, Dr. Junli Gu, Dr. Dongdong Fu and Dr. Kuo-Chin Lien for their insightful discussion and comments during the rebuttal period. 
{\small
\bibliographystyle{ieee_fullname}
\bibliography{egbib}
}

\end{document}